\documentclass[10pt,twocolumn,letterpaper]{article}

\usepackage{cvpr}
\usepackage{times}
\usepackage{epsfig}
\usepackage{graphicx}
\usepackage{amsmath}
\usepackage{amssymb}
\usepackage{booktabs}

\usepackage{makecell}
\usepackage{multirow}
\usepackage{bm}
\usepackage{placeins}

\usepackage{caption}
\usepackage{subfig}
\newcommand{\reffig}[1]{Fig.~\ref{#1}}

\newcommand{\bi}{\begin{itemize}}
\newcommand{\ei}{\end{itemize}}

\newcommand{\ba}{\begin{eqnarray*}}
\newcommand{\ea}{\end{eqnarray*}}

\usepackage[breaklinks=true,bookmarks=false,colorlinks]{hyperref}

\cvprfinalcopy 

\def\cvprPaperID{3115} 

\ifcvprfinal\fi
\begin{document}

\title{DenseReg: Fully Convolutional Dense Shape Regression In-the-Wild}

\author{R{\i}za Alp G\"uler $^{1}$ \hspace{1em}
	George Trigeorgis $^{2}$ \hspace{1em}
	Epameinondas Antonakos $^{2}$\vspace{0.15em}\\
	Patrick Snape $^{2}$\hspace{1em}
    Stefanos Zafeiriou $^{2}$ \hspace{1em}
    Iasonas Kokkinos $^{3}$ \vspace{0.3em}\\
	$^1$INRIA-CentraleSup\'elec, France \hspace{1em} $^2$Imperial College London, UK  \hspace{2em} $^3$University College London, UK\\
	$^{1}${\tt\scriptsize riza.guler@inria.fr}\hspace{1em} 
	$^{2}${\tt\scriptsize\{g.trigeorgis, e.antonakos, p.snape,s.zafeiriou\}@imperial.ac.uk 	}\hspace{1em}	
	$^{3}${\tt\scriptsize i.kokkinos@cs.ucl.ac.uk}
}

\maketitle

\begin{abstract}
In this paper we propose to learn a mapping from image pixels into a dense template  grid through a fully convolutional network.  We formulate this task as a regression problem and train our network by leveraging upon manually annotated facial landmarks ``in-the-wild''. We use such landmarks to establish a dense correspondence field between a three-dimensional object template and the input image, which then serves as the ground-truth for training our regression system. We show that we can combine ideas from semantic segmentation with regression networks, yielding a highly-accurate `quantized regression' architecture.

Our system, called DenseReg, allows us to estimate dense image-to-template correspondences in a fully convolutional manner. As such our network can provide useful correspondence information  as a stand-alone system, while when used as an initialization for Statistical Deformable Models we obtain landmark localization results that largely outperform the current state-of-the-art on the challenging 300W benchmark. 
We thoroughly evaluate our method on a host of facial analysis tasks
and also provide qualitative results for dense human body correspondence. We make our code available at \url{http://alpguler.com/DenseReg.html}  along with supplementary materials.

\end{abstract}
\vspace{-0.5cm}

\section{Introduction}

\begin{figure}[t]
\vspace{-0.05cm}
\begin{center}
  \includegraphics[trim={4.3cm 10cm 0.25cm 4.5cm},clip,width=1 \linewidth ]{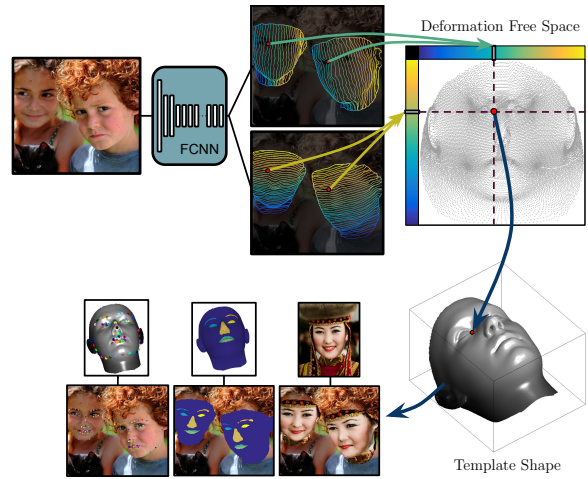}
\end{center}
\vspace{-0.35cm}
   \caption{We introduce a fully convolutional neural network that regresses from the image to a ``canonical'', deformation-free parameterization of the face surface, effectively yielding a dense 2D-to-3D surface correspondence field. Once this correspondence field is available, one can effortlessly solve many image-level problems by backward-warping their canonical solution from the template coordinates to the image domain for the problems of landmark localization, semantic part segmentation, and face transfer.} 
   	
\vspace{-0.35cm}
\label{fig:FaceIntroFig}
\end{figure}

Non-planar object deformations, e.g. due to facial pose or expression, result in challenging but also informative signal variations. Our objective in this paper is to recover this information in a feedforward manner by employing a discriminatively trained convolutional network. Motivated by the gap between discriminatively trained systems for detection and category-level deformable models, we propose a system that combines the merits of both.

In particular, discriminative learning-based approaches typically pursue invariance to shape deformations, for instance by employing  local `max-pooling' operations to ellicit responses that are invariant to {\emph{local}} translations. As such, these models can reliably  detect patterns irrespective of their deformations through efficient, feedforward algorithms. At the same time, however, this discards useful shape-related information and only delivers a single categorical decision per position. Several recent works in deep learning have aimed at enriching deep networks with information about shape  by explicitly modelling {\em the effect} of  similarity transformations \cite{PapandreouKS15}
or non-rigid deformations \cite{JaderbergSZK15,HandaBPSMD16,ChenHW016}; several of these have found success in classification \cite{PapandreouKS15}, fine-grained recognition  \cite{JaderbergSZK15}, and also face detection \cite{ChenHW016}. There are works \cite{lades1993distortion,pedersoli2015elastic} that model the deformation via optimization procedures, whereas we obtain it in a feedforward manner and in a single shot. In these works, shape is treated as a nuisance, while we treat it as the goal in itself. Recent works on 3D surface correspondence \cite{ Br1,Br2} have shown the merit of CNN-based unary terms for correspondence. 
There are works that address the problem of  establishing dense correspondence for the human body from static RGBD images\cite{taylor2012vitruvian,pons2015metric,wei2016dense}.
In our case we tackle the much more challenging task of establishing a 2D to 3D correspondence {\emph{in the wild}} by leveraging upon recent advances in semantic segmentation \cite{CP2015Semantic}. To the best of our knowledge, the  task of explicitly recovering dense correspondence in the wild has not been addressed yet in the context of deep learning. 

By contrast, approaches that rely on  Statistical Deformabe Models (SDMs), such as Active Appearance Models (AAMs) or 3D Morphable Models (3DMMs) aim at explicitly recovering  dense correspondences between a deformation-free template and the observed image, rather than trying to discard them. This allows to both represent shape-related information (\textit{e.g.} for facial expression analysis) and also to obtain invariant decisions after registration (\textit{e.g.} for identification). Explicitly representing shape  can have substantial performance benefits, as is witnessed in the majority of facial analysis tasks requiring detailed face information e.g.  landmark localisation \cite{sagonas2016300}, 3D pose estimation, as well as 3D face reconstruction ``in-the-wild'' \cite{jourabloo2016large}, where SDMs consitute the current state of the art. 

 However SDM-based methods are limited in two respects. Firstly they require an initialization from external systems, which can become increasingly challenging for elaborate SDMs: both AAMs and 3DMMs require at least a bounding box as initialization and 3DMMs may further require position of specific facial landmarks. Furthermore, SDM fitting  requires iterative, time-demanding optimization algorithms, especially when the initialisation is far from the solution. The advent of Deep Learning has made it possible to replace the iterative optimization task with iterative regression problems \cite{trigeorgis2016mnemonic}, but this does not alleviate the need for initialization and multiple iterations.

In this work we aim at bridging these two approaches, and introduce a discriminatively trained network to obtain, in a fully-convolutional manner, dense correspondences between an input image and a deformation-free template coordinate system. 

In particular, we exploit the availability of manual facial landmark annotations ``in-the-wild'' in order to fit a 3D template; this provides us with a dense correspondence field, from the image domain to the 2-dimensional, $U-V$ parameterization of the face surface. We then train a fully convolutional network that densely regresses from the image pixels to this $U-V$ coordinate space.

This provides us with dense and fine-grained correspondence information, as in the case of SDMs, while at the same time being independent of any initialization procedure, as in the case of discriminatively trained `fully-convolutional' networks. We demonstrate that the performance of certain tasks, such as facial landmark localisation or segmantic part segmentation, is largely improved by using the proposed network. 

Even though the methodology is general, this paper is mainly concerned with human faces.
The architecture for the case of human face is described in  Fig.~\ref{fig:FaceIntroFig}.

Our approach can be seen in two complementary manners: first, it provides a stand-alone, feedforward alternative to the combination of initialization with iterative fitting typically used in SDMs. This allows us to have a feedforward system that solves both the detection and correspondence problems at approximate $7-8$ frames per second for a $300\times300$ input image.
Secondly, our approach can also be understood as an initialization procedure for SDMs which gets them started from a much more accurate position than the bounding box, or landmark-based initializations currently employed in the face analysis literature. When taking this approach we observe substantial gains over the current state-of-the-art systems. 

We can summarize our contributions as follows:
\begin{itemize}
\item We introduce the task of dense shape regression in the setting of CNNs, and exploit the SDM-based notion of a deformation-free UV-space to construct target ground-truth signals (Sec.\ref{sec:SDMs}).
\item We propose a carefully-designed fully-convolutional shape regression system that exploits ideas from semantic segmentation and dense regression networks. Our \textit{quantized regression} architecture~(Sec.\ref{sec:quantized}) is shown to substantially outperform simpler baselines that consider the task as a plain regression problem. 
\item We use dense shape regression to jointly tackle a multitude of problems, such as landmark localization or semantic segmentation.

In particular, the template coordinates allow us to `copy' multiple annotations constructed on a single template system, and thereby tackle multiple problems in a single go.
\item We use the regressed shape coordinates for the initialization of SDMs; systematic evaluations on facial analysis benchmarks show that this yields substantial performance improvements  on tasks ranging from landmark localization to semantic segmentation.
\item We  demonstrate the generic nature of the method by applying it to the task of estimating dense correspondences for other deformable surfaces, such as the human body and the human ear. 
\end{itemize}

\section{From SDMs to Dense Shape Regression}
\label{sec:SDMs}

Following the deformable template paradigm \cite{yuille1991deformable,Grenander1991}, we consider that object instances are obtained by deforming a prototypical object, or `template', through  dense deformation fields. 
This makes it possible  to factor  object variability within a category into variations that are associated to  deformations, generally linked to the object's 2D/3D shape, and variations that are associated to appearance (or, `texture' in graphics), e.g. due to facial hair, skin color, or illumination. 

This factorization largely simplifies the  modelling task. SDMs use it as a stepping stone for the construction of parametric models of deformation and appearance. For instance, in AAMs a combination of Procrustes Analysis, Thin-Plate Spline warping and PCA is the standard pipeline for learning a low-dimensional linear subspace that captures category-specific shape variability \cite{cootes2001active}. Even though we have a common starting point, rather than trying to construct a linear generative model of deformations, we treat the image-to-template correspondence as a vector field that our network tries to regress.

\begin{figure}[!t]
\centering
\includegraphics[trim={7.2cm 18.3cm 2.7cm 6.5cm}, clip, width=\linewidth]{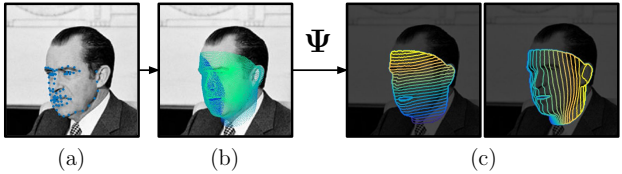}
\caption{Ground-truth generation: \emph{(a)} Annotated landmarks. \emph{(b)} Template shape morphed based on the landmarks. \emph{(c)} Deformation-free coordinates (${u^h}$ and ${u^v}$), obtained by unwrapping the template shape, transferred to image domain. }
\vspace{-0.35cm}
\label{fig:GT}
\end{figure}

In particular, we start from a template $\bm{X} = [\bm{x}_1^\top ,\bm{x}_2^\top,...\bm{x}_m^\top]^\top \in \mathbb{R}$, where each $\bm{x}_j \in \mathbb{R}^3$ is a vertex location of the mesh in 3D space. 

This template could be any 3D facial mesh, but in practice it is most useful to use a topology that is in correspondence with a 3D statistical shape model such as \cite{booth3d} or \cite{paysan20093d}.
We compute a bijective mapping $\psi$, from template mesh $\bm{X}$ to the 2D canonical space $\bm{U} \in \mathbb{R}^{2\times m}$, such that  
\begin{equation}
\psi(\bm{x}_j) \mapsto \bm{u}_j \in \bm{U}  \quad  ,  \quad  \psi^{-1}(\bm{u}_j) \mapsto \bm{x}_j .
\end{equation} 
The mapping $\psi$ is obtained via the cylindrical unwrapping described in \cite{booth2014optimal}. Thanks to the cylindrical unwrapping, we can interpret these coordinates as being the horizontal and vertical coordinates while moving on the face surface: ${u}_j^h \in [0,1]$ and ${u}_j^v \in [0,1]$. Note that this semantically meaningful parameterization has no effect on the operation of our method.

We exploit the availability of landmark annotations ``in the wild'', to fit the template face to the image by obtaining a coordinate transformation for each vertex $\bm{x}_j$. 
We use the fittings provided by \cite{zhu2016face} which were fit using a modified 3DMM implementation \cite{romdhani2005estimating}. However, for the purpose of this paper, we require a per-pixel estimate of the location in UV space on our template mesh and thus do not require an estimate of the projection or model parameters as required by other 3D landmark recovery methods \cite{jourabloo2016large,zhu2016face}. The per-pixel UV coordinates are obtained through rasterization of the fitted mesh and non-visible vertices are culled via z-buffering.

As  illustrated in \reffig{fig:GT}, once the transformation from the template face vertices to the morphed vertices is established, the  $\bm{u}_j$ coordinates of each visible vertex on the canonical face can be transferred to the image space. This establishes the ground truth signal for our subsequent regression task.

\begin{figure*}[t]
\centering
\includegraphics[trim={5.2cm 20cm 3.7cm 4.5cm}, clip, width=0.95\linewidth ]{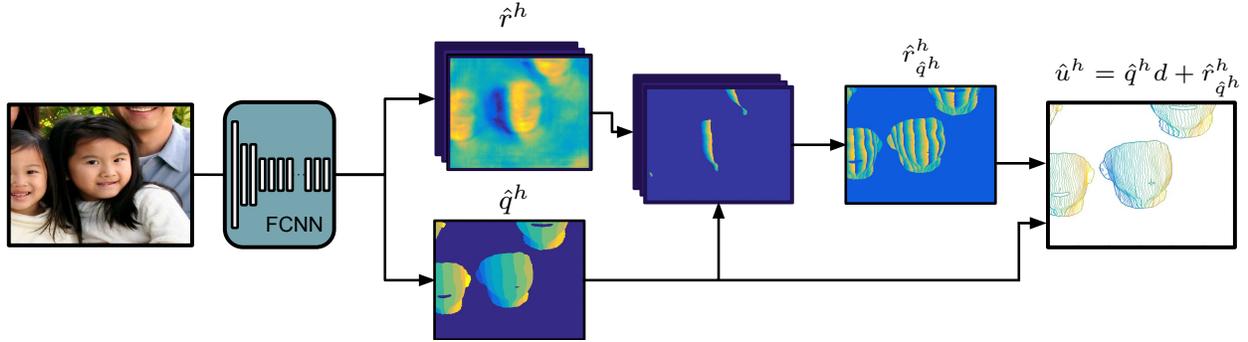}
\caption{Proposed Quantized Regression Approach for the horizontal correspondence signal: The continuous  signal is regressed by first estimating a grossly quantized (or, discretized) function  through a classification branch. For each quantized value $\hat{q}^h$ we use a separate residual regression unit's prediction, $\hat{r}^h_{\hat{q}^h}$, effectively multiplexing the different residual predictions. These are added to the quantized prediction, yielding a smooth and accurate correspondence field. }
\vspace{-0.35cm}
\label{fig:Pipeline}
\end{figure*}

\section{Fully Convolutional Dense Shape Regression }
\label{sec:quantized}%

	Having described how we establish our supervision signal, we now turn to the task of estimating it through a convolutional neural network (CNN). 
		Our aim is to estimate at any image pixel that belongs to a face region the values of  $\bm{u} =[u^h, u^v]$. We need to also identify non-face pixels,  e.g. by predicting a `dummy' output. 
		
	One can phrase this problem as a generic regression task and attack it with the powerful machinery of CNNs. Unfortunately, the best performance that we could obtain this way was quite underwhelming, apparently due to the task's complexity. Our approach is to quantize and estimate the quantization error separately for each quantized value. Instead of directly regressing $u$, the quantized regression approach lets us solve a set of easier sub-problems, yielding improved regression results.

In particular,	instead of  using a CNN as a `black box' regressor, we draw inspiration from the success of recent works on semantic part  segmentation \cite{tsogkas2015deep,CP2016Deeplab}, and landmark classification \cite{bulat2016human,bulat2016two}. These works have shown that CNNs can deliver remarkably accurate predictions when trained to predict \textit{categorical variables}, indicating for instance the facial part or landmark corresponding to each pixel. 
	
	Building on these successes, we propose a hybrid method that combines a classification with a regression problem. Intuitively, we first identify a coarser face region that can contain each pixel, and then obtain a refined, region-specific prediction of the pixel's $U-V$  field. As we will describe below, this yields substantial gains in performance when compared to the baseline of a generic regression system. 
	
	We identify  facial regions  by using a simple geometric approach.
	We tesselate the template's surface with a cartesian grid, by uniformly and separately quantizing the $u^h$ and $u^v$ coordinates into $K$ bins, where $K$ is a design parameter. For any image that is brought into correspondence with the template domain, this induces a discrete labelling, which can be recovered by training a  CNN for classification.
	
\begin{figure}[b]
\begin{center}
   \includegraphics[width=1\linewidth ]{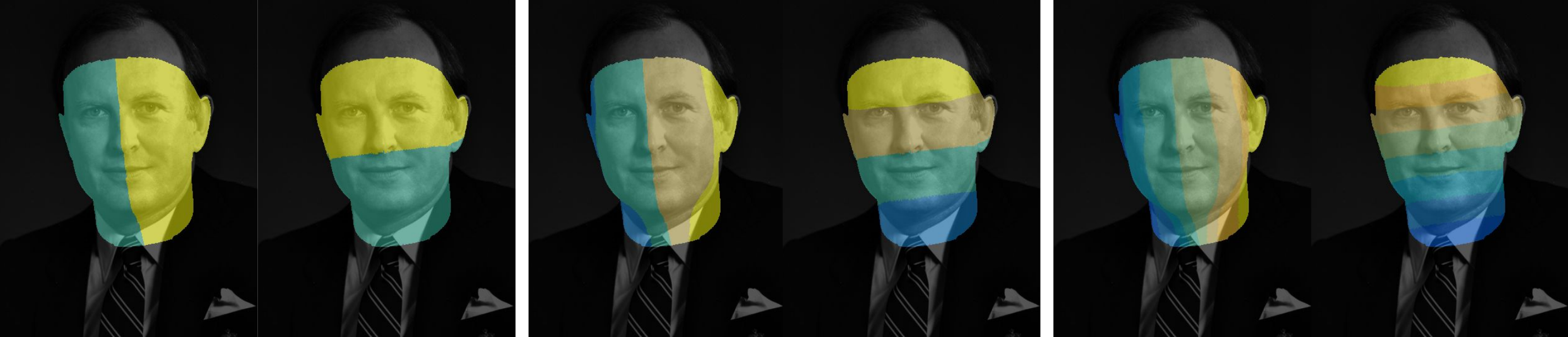}
\end{center}
   \caption{Horizontal and vertical tesselations obtained using $K=2,4$ and $8$ bins.}
   \vspace{-0.5cm}
\label{fig:DiscreteFaces}
\end{figure}

	 On Fig.~\ref{fig:DiscreteFaces}, the tesselations of different granularities are visualized. For a sufficiently large value of $K$ even a plain classification result could provide a reasonable estimate of the pixel's correspondence field, albeit with some staircasing effects. The challenge here is that as the granularity of these discrete labels becomes increasingly large, the amount of available training data decreases and label complexity increases. A more detailed analysis on the effect of label-space granularity to segmentation performance is provided in supplementary materials.
	
	We propose to combine powerful classification results with a regression problem that will yield a  refined  correspondence estimate. For this, we compute the residual between the desired and quantized $U-V$  coordinates and add a separate module that tries to regress it. We train a separate regressor per facial region, and at any pixel only penalize the regressor loss for the responsible face region. We can interpret this form as a `hard' version of a mixture of regression experts \cite{JordanJ94}. This interpretation is further elaborated upon in the supplementary material.
	
		The horizontal and vertical components $u^h,u^v$ of the correspondence field are predicted separately. This results in a substantial reduction in computational and sample complexity -  For $K$ distinct U and V bins we have $K^2$ regions; the classification is obtained by combining 2 $K$-way classifiers. Similarily, the regression mapping involves $K^2$ regions, but only uses $2 K$ one-dimensional regression units. The pipeline for quantized face shape regression is provided in Fig.~\ref{fig:Pipeline}.

	We now detail the training and testing of this network;  for simplicity we only describe the horizontal component of the mapping. 
	From the ground truth construction, every position $\bm{x}$ is associated with a scalar ground-truth value $u^h$. Rather than trying to predict $u^h$ as is, we transform it into a pair of discrete $q^h$ and continuous $r^h$ values, encoding the quantization and residual respectively:
	\begin{equation} 
	q^h =  \lfloor {\frac{u^h}{d}} \rfloor, \quad  r_i^h =   \left(u^h_i - q^h_i d  \right),
	\end{equation}
	where $d = \frac{1}{K}$ is the quantization step size (we consider $u^h,u^v$ coordinates to lie in $[0,1$]).
	
	Given a common CNN trunk, we use two classification branches to predict $q^h, q^v$ and two regression branches to predict $r^h,r^v$ as convolution layers with kernel size $1\times1$. As mentioned earlier, we employ separate regression functions per region, which means that at any position we have $K$ estimates of the horizontal residual vector, $\hat{r}^h_{i},~i=1,\ldots,K$.
	
	At test time, we let the network predict the discrete bin $\hat{q}^h$ associated with every input position, and then use the respective regressor output $\hat{r}^h_{\hat{q}^h}$ to obtain an estimate of $u$:
	\begin{equation}   
	\hat{u}^h =  \hat{q}^h d + \hat{r}^h_{\hat{q}^h}
	\end{equation}

For the $q^h$ and $q^v$, which are modeled as categorical distributions,  we use  softmax followed by the cross entropy loss. For estimating $\hat{r}^h$ and $\hat{r}^v$, we use a normalized version of the smooth $L_1$ loss~\cite{girshick2015fast}. The normalization is obtained by dividing the loss by the number of pixels that contribute to the loss.



\begin{figure}[!htb]
    \centering
    \begin{minipage}{.7\linewidth}
        \centering
        \includegraphics[trim={0.5cm 0.5cm 0.7cm 0.5cm},clip,width=0.9\textwidth]{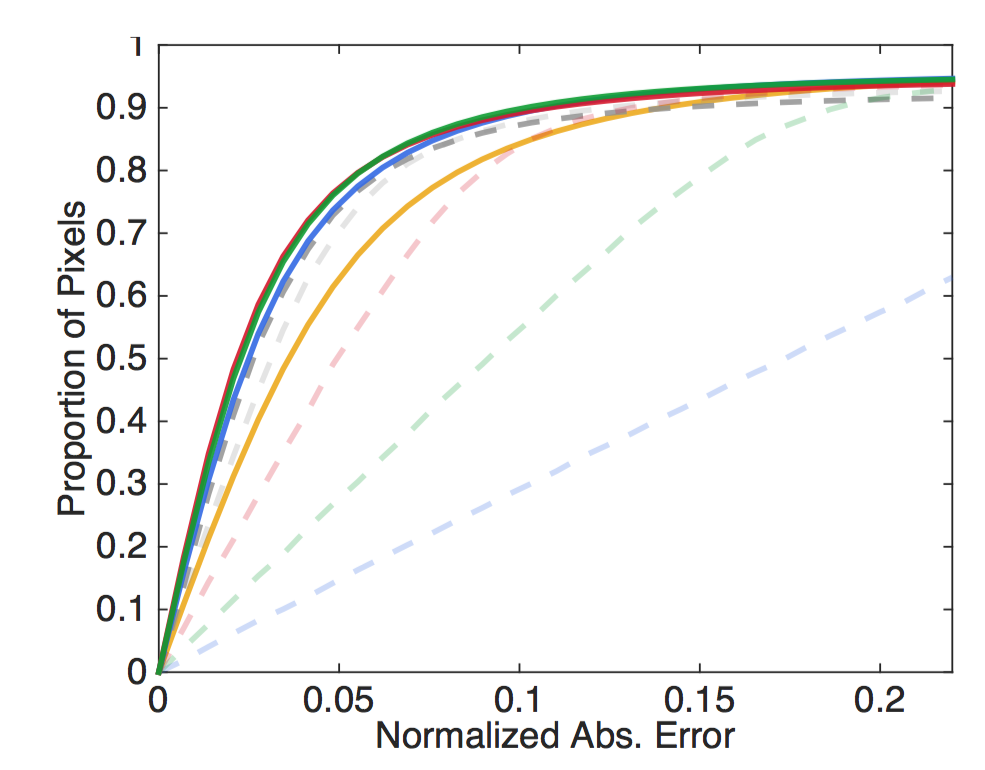}
    \end{minipage}%
    \begin{minipage}{0.3\linewidth}
        \centering
        \includegraphics[trim={1.5cm 1cm 10cm 1cm},clip,width=1\textwidth]{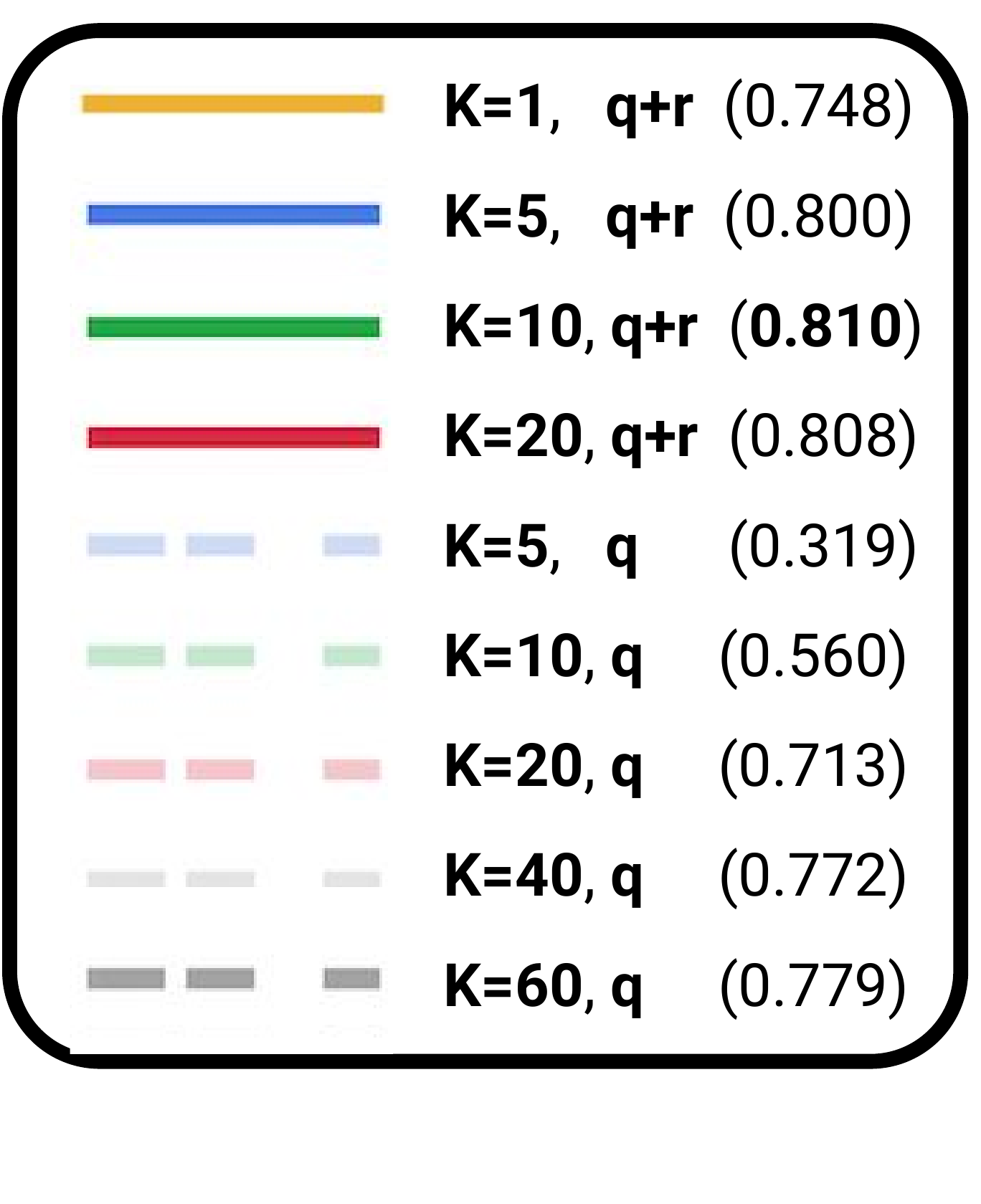}
    \end{minipage}
    \vspace{-0.35cm}
    \caption{Performance of $q$ and $r$, branches for various tesselation granularities, $K$. Areas under the curve(AUC) are reported.}
    \vspace{-0.15cm}
    \label{fig:exp}
    \vspace{-0.15cm}

\end{figure}

Compared to plain regression of the coordinates, the proposed method achieves much better results. 
In Fig.\ref{fig:exp} we report results of an experiment that evaluates the contribution of the q-r branches separately for different granularities. The results for the quantized branch are evaluated by transforming the discrete horzintal/vertical label into the center of the region corresponding to the quantized horizontal/vertical value respectively.  The results  show the merit of adopting the classification branch, as the finely quantized results(K=40,60) yield  better coordinate estimates with respect to the non-quantized alternative {(K=1)}. After K=40, we observe an increase in the failure rate for the quantized branch. The experiment reveals that the proposed quantized regression outperforms both \textit{non-quantized} and the best of \textit{only-quantized} alternatives. 

\section{Experiments}
\label{sec:experiments}
Herein, we evaluate the performance of the proposed method (referred to as \texttt{DenseReg}) on various tasks. 
In the following sections, we first describe the training setup (Sec.~\ref{sec:training_setup}) and then present extensive quantitative and qualitative results on \emph{(i)}~semantic segmentation (Sec.~\ref{sec:exp_semantic_segmentation}), \emph{(ii)}~landmark localization on static images (Sec.~\ref{sec:exp_landmark_localization}), \emph{(iii)}~deformable tracking (Sec.~\ref{sec:exp_deformable_tracking}),  \emph{(iv)}~ monocular depth estimation (Sec.~\ref{sec:exp_depth}), \emph{(v)}~ dense correspondence on human bodies (Sec.~\ref{sec:exp_human}), and \emph{(vi)}~human ear landmark localization (Sec.~\ref{sec:exp_ear}).

\subsection{Training Setup}
\label{sec:training_setup}

\textbf{Training Databases.} We train our system using the 3DDFA data of \cite{zhu2016face}. The 3DDFA data provides projection and 3DMM model parameters for the Basel \cite{paysan20093d} + FaceWarehouse \cite{cao2014facewarehouse} model for each image of the 300W database. We use the topology defined by this model to define our UV space and rasterize the images to obtain per-pixel ground truth UV coordinates.  Our training set consists of the LFPW trainset, Helen trainset and AFW, thus 3148 images
that are captured under completely unconstrained conditions
and exhibit large variations in pose, expression, illumination,
age, etc.
 Many of these images contain multiple faces, some of which are not annotated. We deal with this issue by employing the out-of-the-box DPM face detector of Mathias et al.~\cite{mathias2014face} to obtain the regions that contain a face for all of the images. The detected regions that do not overlap with the ground truth landmarks do not contribute to the loss. For training and testing, we have rescaled the images such that their largest side is 800 pixels.

\textbf{CNN Training.} For the dense regression network, we adopt a ResNet101~\cite{He2015} architecture with dilated convolutions (atrous)~\cite{CP2015Semantic,mallat1999wavelet}, such that the stride of the CNN is $8$. We use bilinear interpolation to upscale both the $\hat{q}$ and $\hat{r}$ branches before the losses. The losses are applied at the input image scale and back-propagated through interpolation. We apply a weight to the smooth $L1$ loss layers to balance their contribution. In our experiments, we have used a weight of $40$ for quantized~($d=0.1$) and a weight of $70$ for non-quantized regression, which are determined by a coarse cross validation. We initialize the training with a network pre-trained for the MS COCO segmentation task~\cite{lin2014microsoft}. The new layers are initialized with random weights drawn from Gaussian distributions. Large weights of the regression losses can be problematic at initialization even with moderate learning rates. To cope with this, we use initial training with a lower learning rate for a \textit{warm start} for a few iterations. We then use a base learning rate of $0.001$ with a polynomial decay policy for $20k$ iterations with a batch size of $10$ images. During training, each sample is randomly scaled with one of the ratios $[0.5, 0.75, 1, 1.25, 1.5]$ and cropped to form a fixed $321\times321$ input image.

\subsection{Semantic Segmentation}
\label{sec:exp_semantic_segmentation}


As discussed in Sec.~\ref{sec:SDMs}, any labelling function defined on the template shape can be transferred to the image domain using the regressed coordinates. One application that can be naturally represented on the template shape is semantic segmentation of facial parts. 
To this end, we manually defined a segmentation mask of $8$ classes  (right/left eye, right/left eyebrow, upper/lower lip, nose, other) on the template shape, as shown in Fig.~\ref{fig:FaceIntroFig}. 
\begin{figure}[h]
\vspace{-0.25cm}
\centering
\includegraphics[width=\linewidth ]{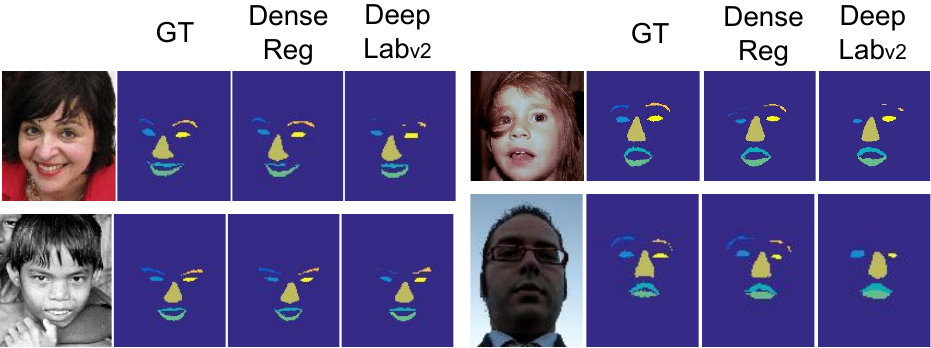}
\caption{Exemplar semantic segmentation results.}
\label{fig:Semantic}
\vspace{-0.25cm}

\end{figure}

We compare against a state-of-the-art semantic part segmentation system (DeepLab-v2)~\cite{CP2016Deeplab} which is based on the same ResNet-101 architecture as our proposed DenseReg. We train DeepLab-v2 on the same training images (i.e. LFPW trainset, Helen trainset and AFW).  We generate the ground-truth segmentation labels for both training and testing images by transferring the segmentation mask using the ground-truth deformation-free coordinates explained in Sec.~\ref{sec:SDMs}. We employ the Helen testset~\cite{le2012interactive} for the evaluation.

Table~\ref{tab:Semantic} reports evaluation results using the intersection-over-union (IoU) ratio. Additionally, Fig.~\ref{fig:Semantic} shows some qualitative results for both methods, along with the ground-truth segmentation labels. The results indicate that the DenseReg outperforms DeepLab-v2. The reported improvement is substantial for several parts, such as eyebrows and lips. We believe that this result is significant given that DenseReg is not optimized for the specific task-at-hand, as opposed to DeepLab-v2 which was trained for semantic segmentation. This performance difference can be justified by the fact that DenseReg was exposed to a  richer label structure during training, which reflects the underlying variability and structure of the problem. 


\begin{table}[h]
\centering
\scalebox{0.9}{
\begin{tabular}{|l|cc|}
\hline
\multirow{2}{*}{\emph{Class}} & \multicolumn{2}{c|}{\emph{Methods}}\\
\cline{2-3}
 & \textbf{DenseReg}  & Deeplab-v2 \\
\hline\hline
Left Eyebrow     & 48.35 & 40.57\\
Right Eyebrow    & 46.89 & 41.85\\
Left Eye         & 75.06 & 73.65\\
Right Eye        & 73.53 & 73.67\\
Upper Lip        & 69.52 & 62.04\\
Lower Lip        & 75.18 & 70.71\\
Nose             & 87.71 & 86.76\\
Other            & 99.44 & 99.37\\
\hline\hline
Average          & \textbf{71.96} & 68.58\\
\hline
\end{tabular}
}
\caption{Semantic segmentation accuracy on Helen testset measured using intersection-over-union (IoU) ratio.}
\vspace{-0.05cm}
\label{tab:Semantic}
\end{table}




\subsection{Landmark Localization on Static Images}
\label{sec:exp_landmark_localization}

\begin{figure}[!b]
\vspace{-0.25cm}
\centering
\includegraphics[width=0.9\linewidth]{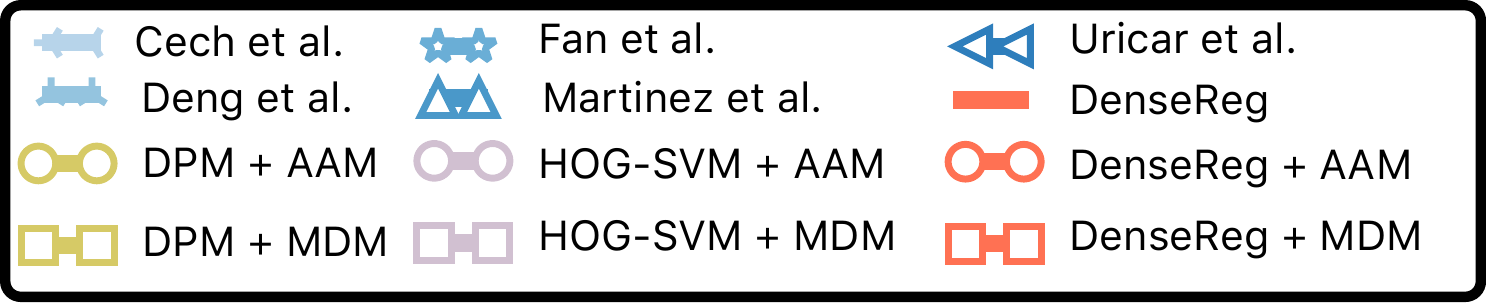}
\includegraphics[width=0.9\linewidth]{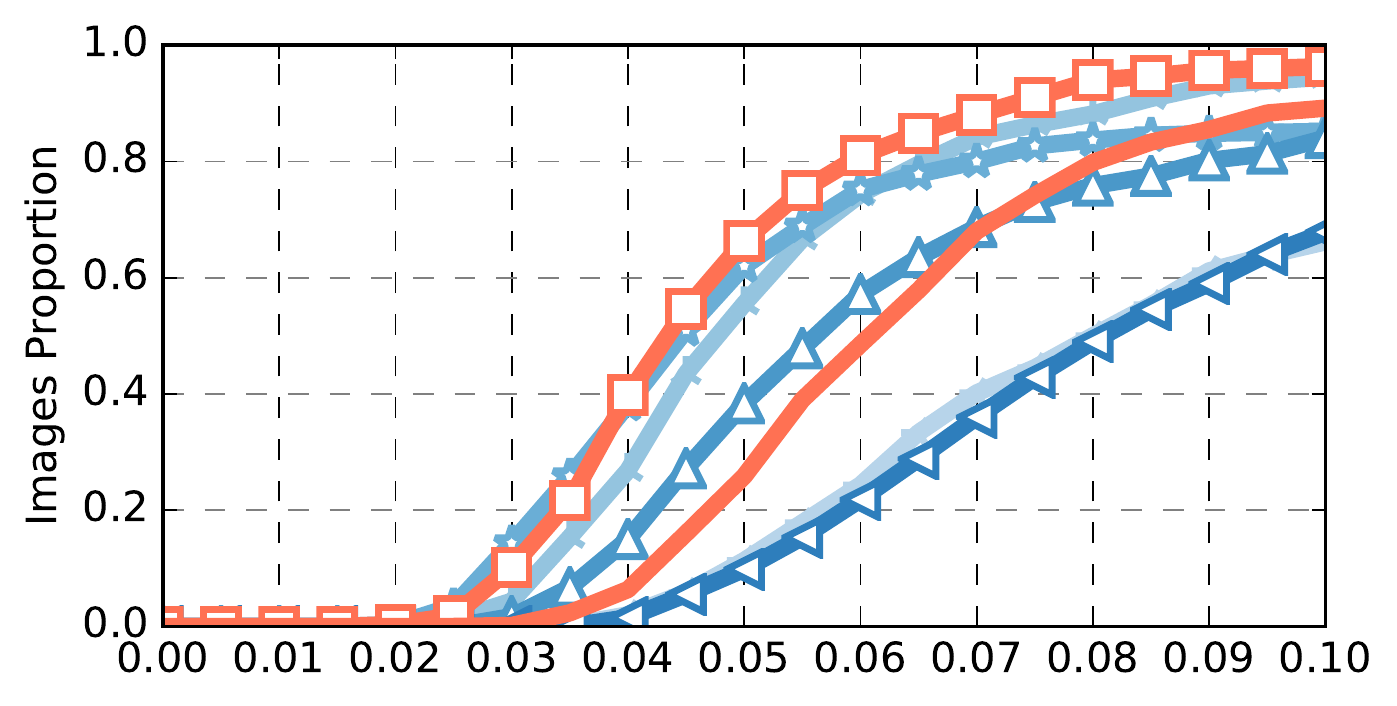}
\includegraphics[width=0.9\linewidth]{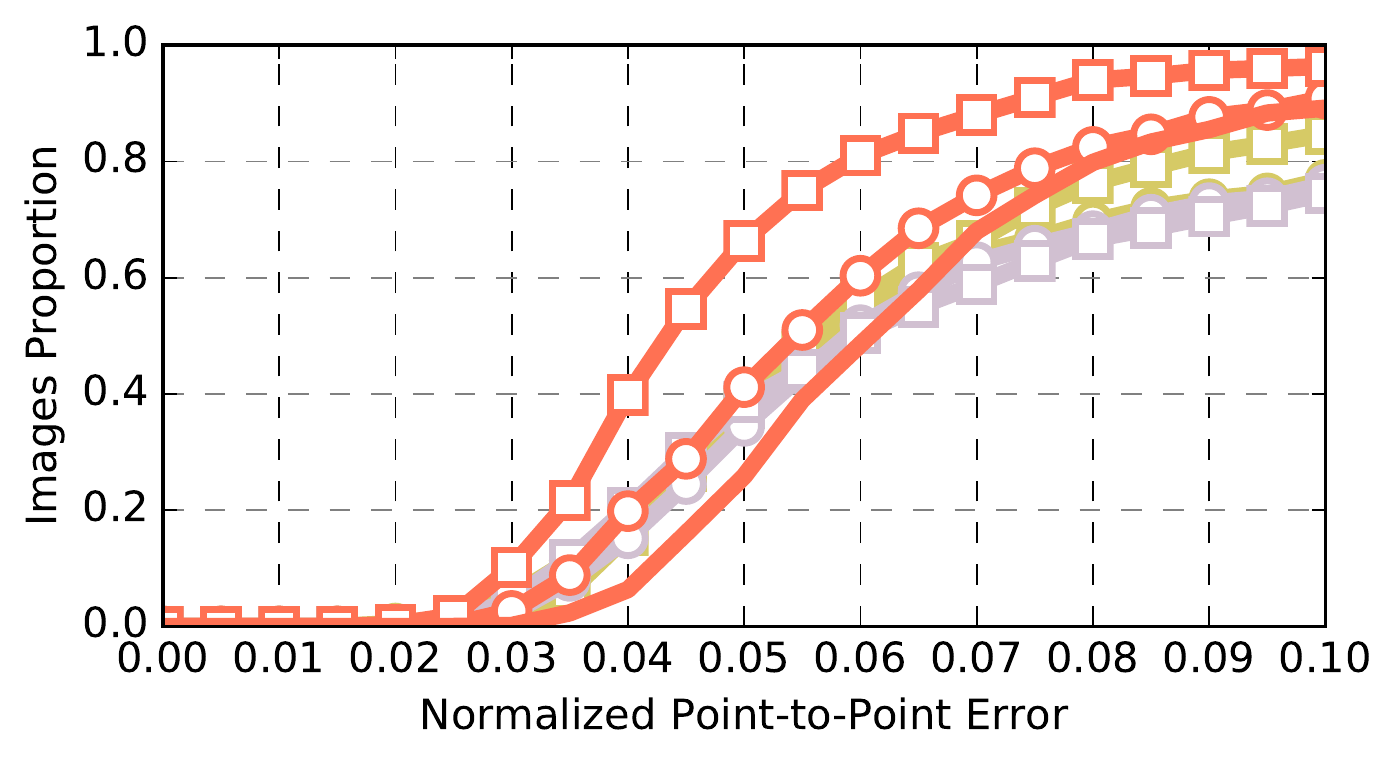}
\caption{Landmark localization results on the 300W testing dataset using 68 points. Accuracy is reported as Cumulative Error Distribution of RMS point-to-point error normalized with interocular distance. \emph{Top:} Comparison with state-of-the-art. \emph{Bottom:} Self-evaluation results.}
\vspace{-0.15cm}
\label{fig:300w}
\end{figure}

DenseReg can be readily used for the task of facial landmark localization on static images. Given the landmarks' locations on the template shape, it is straightforward to estimate the closest points in the deformation-free coordinates on the images. The local minima of the Euclidean distance between the estimated coordinates and the landmark coordinates are considered as detected landmarks. In order to find the local minima, we simply analyze the connected components separately. Even though more sophisticated methods for covering ``touching shapes'' can be used, we found that this simplistic approach is sufficient for the task. 

\begin{figure*}[h]
\centering

\subfloat{\includegraphics[width=0.13775\textwidth]{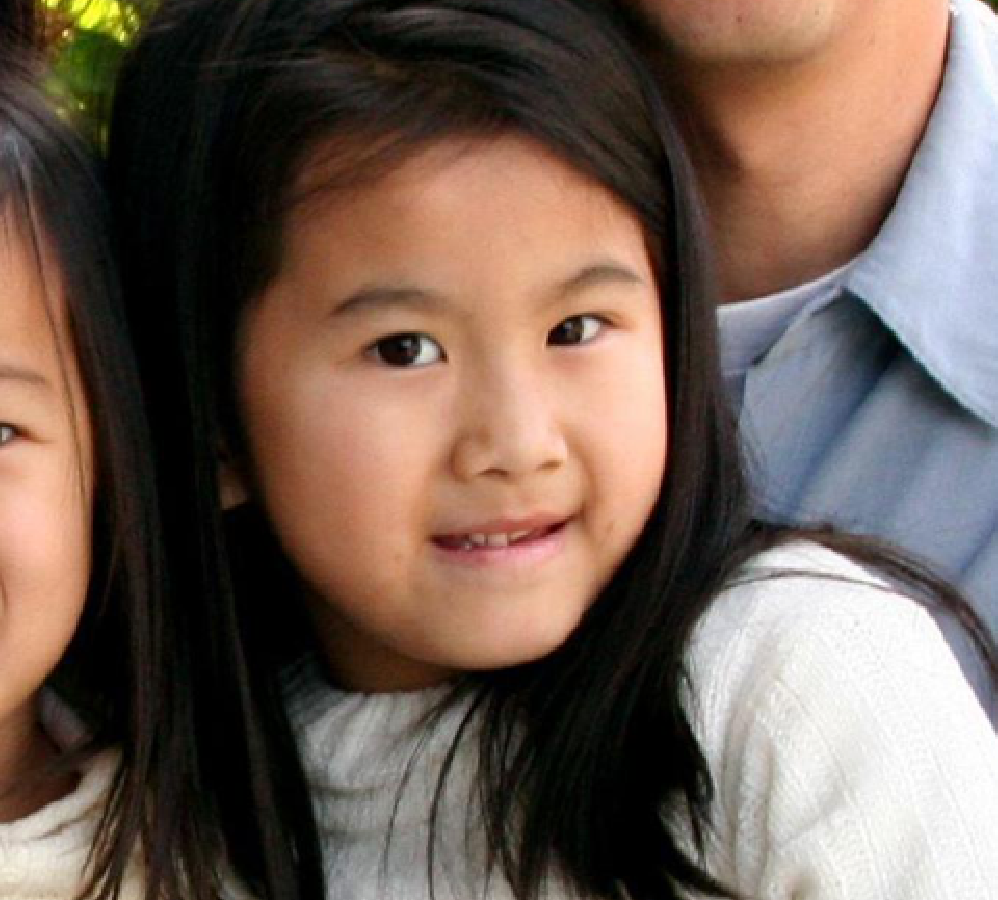}}\hspace{0.0005cm}
\subfloat{\includegraphics[width=0.13775\textwidth]{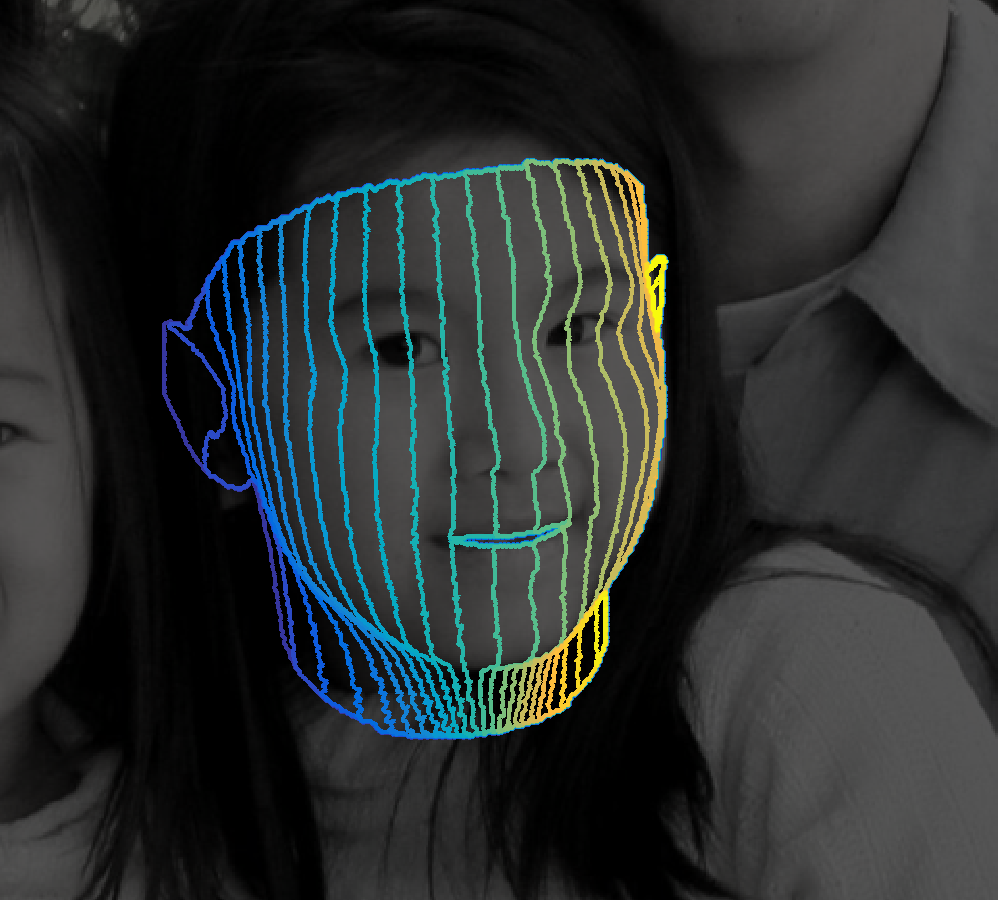}}\hspace{0.0005cm}
\subfloat{\includegraphics[width=0.13775\textwidth]{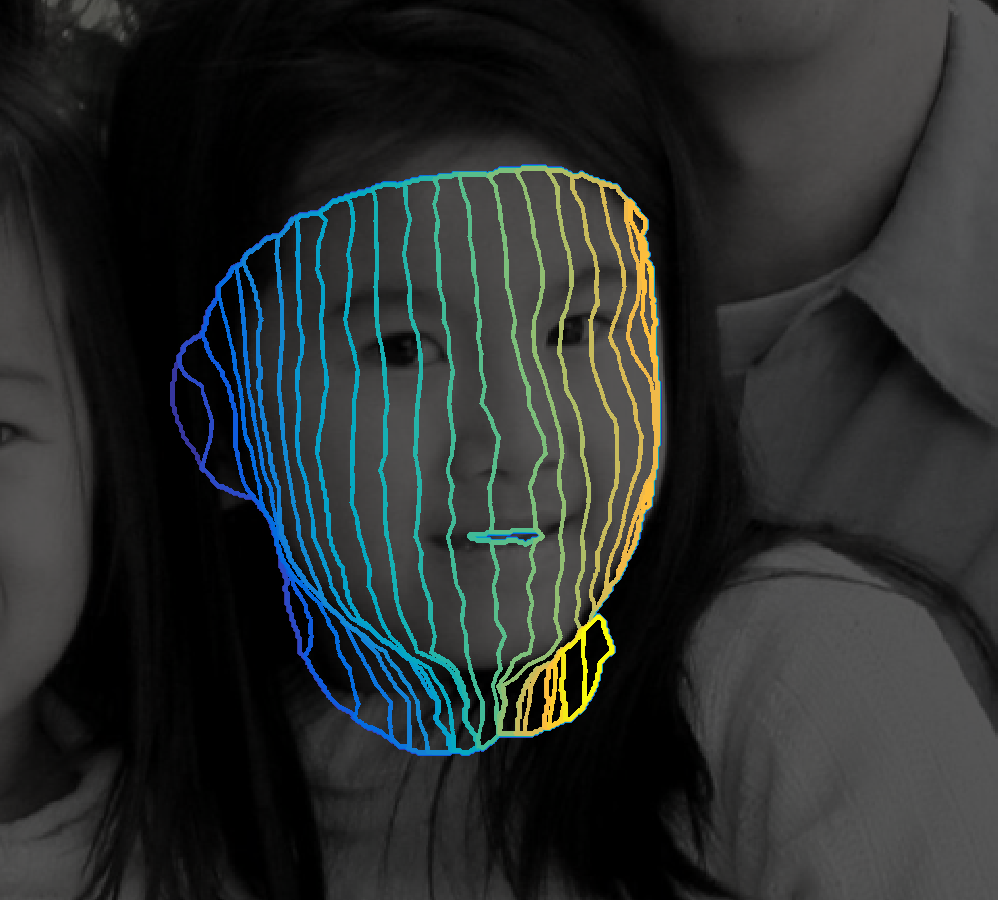}}\hspace{0.0005cm}
\subfloat{\includegraphics[width=0.13775\textwidth]{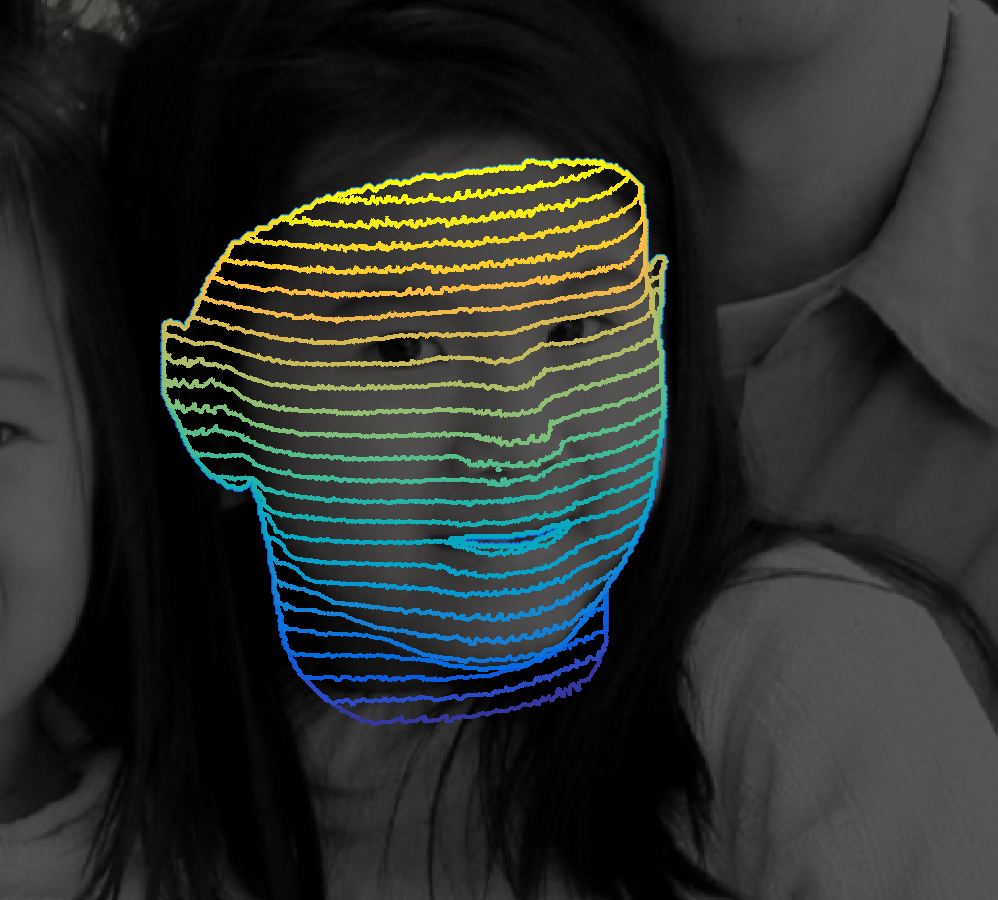}}\hspace{0.0005cm}
\subfloat{\includegraphics[width=0.13775\textwidth]{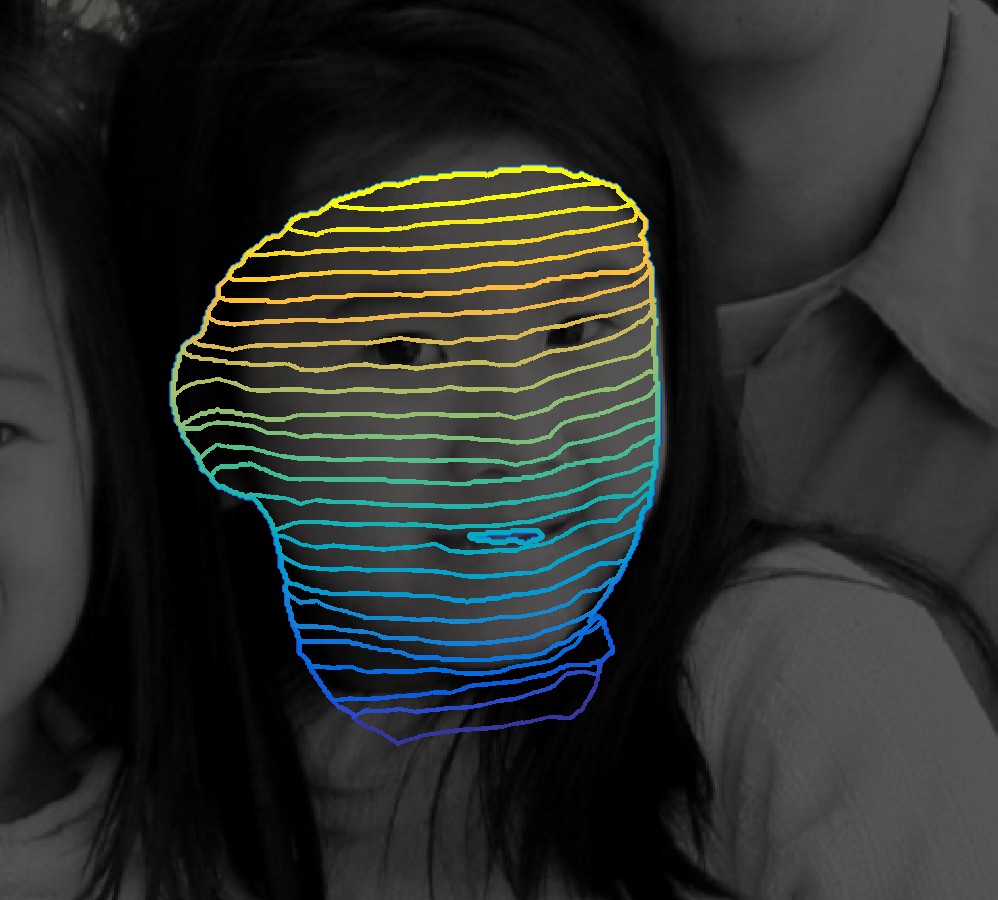}}\hspace{0.0005cm}
\subfloat{\includegraphics[width=0.13775\textwidth]{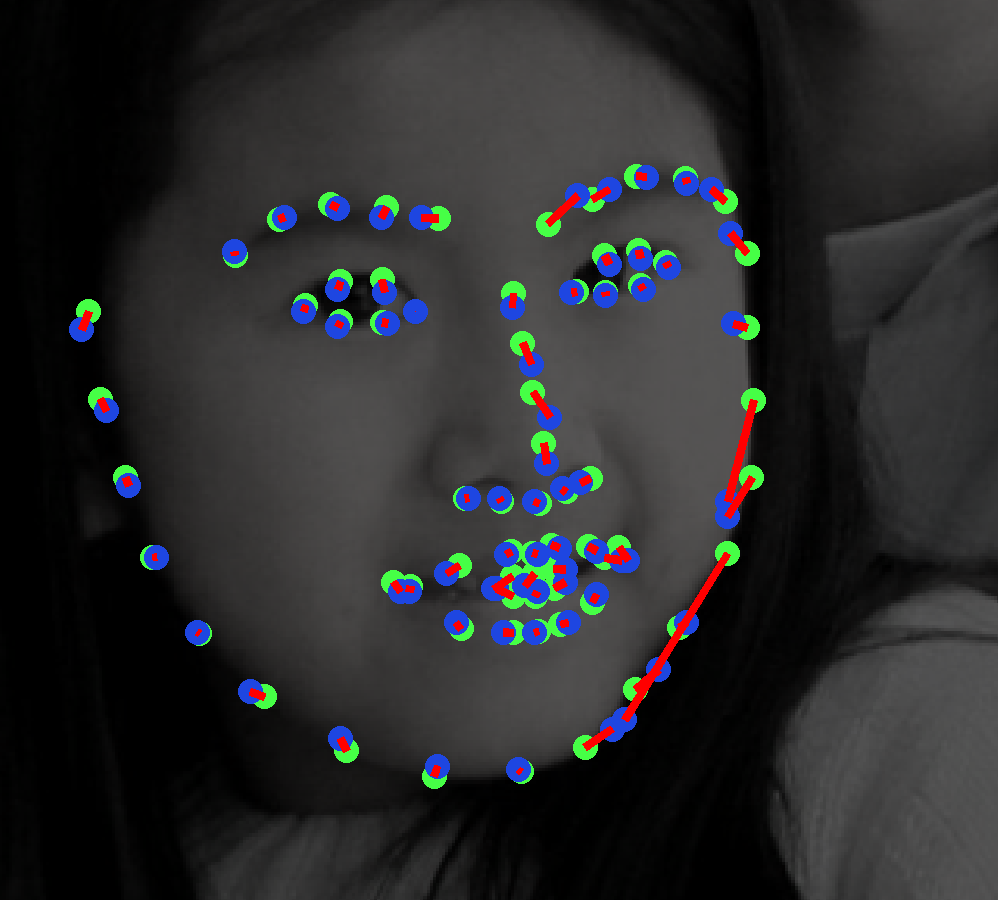}}\hspace{0.0005cm}
\subfloat{\includegraphics[width=0.13775\textwidth]{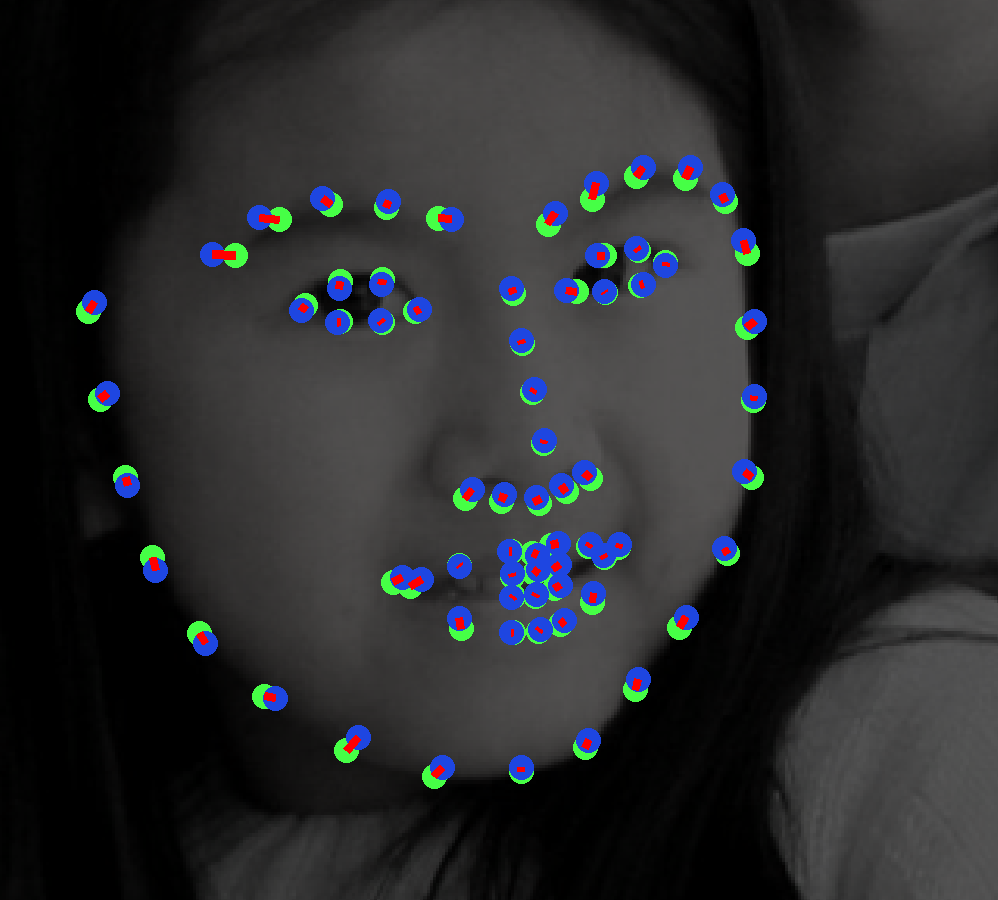}}\\
\vspace{-0.3cm}
\subfloat{\includegraphics[width=0.13775\textwidth]{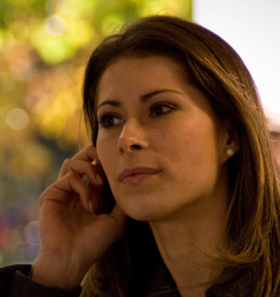}}\hspace{0.0005cm}
\subfloat{\includegraphics[width=0.13775\textwidth]{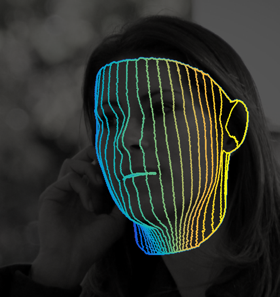}}\hspace{0.0005cm}
\subfloat{\includegraphics[width=0.13775\textwidth]{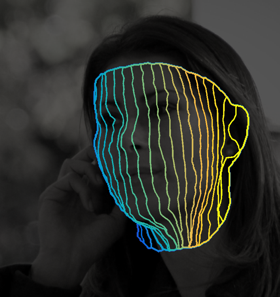}}\hspace{0.0005cm}
\subfloat{\includegraphics[width=0.13775\textwidth]{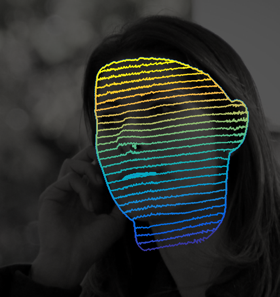}}\hspace{0.0005cm}
\subfloat{\includegraphics[width=0.13775\textwidth]{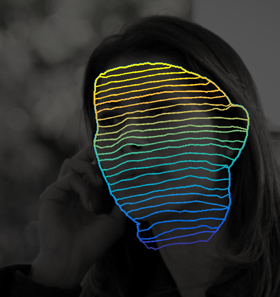}}\hspace{0.0005cm}
\subfloat{\includegraphics[width=0.13775\textwidth]{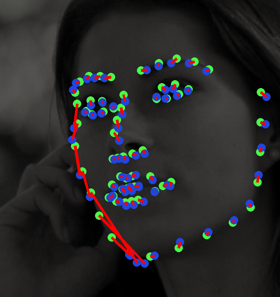}}\hspace{0.0005cm}
\subfloat{\includegraphics[width=0.13775\textwidth]{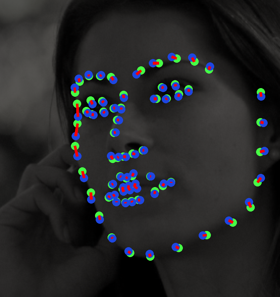}}\\
\vspace{-0.3cm}
\subfloat[Input Image]{\includegraphics[width=0.13775\textwidth]{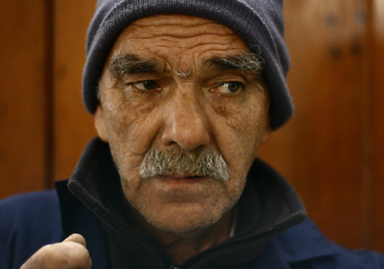}}\hspace{0.0005cm}
\subfloat[Groundtruth $U$ ($u^h$) ]{\includegraphics[width=0.13775\textwidth]{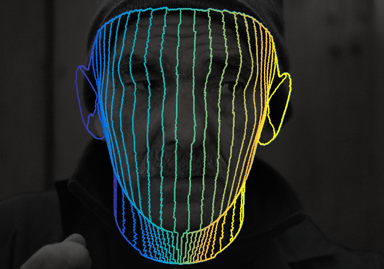}}\hspace{0.0005cm}
\subfloat[Estimated $U$ ($\hat{u}^h$) ]{\includegraphics[width=0.13775\textwidth]{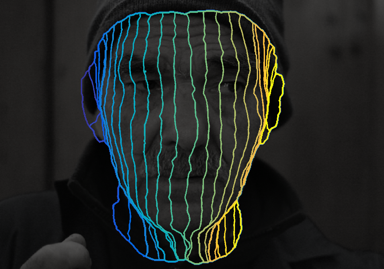}}\hspace{0.0005cm}
\subfloat[Groundtruth $V$ ($u^v$)]{\includegraphics[width=0.13775\textwidth]{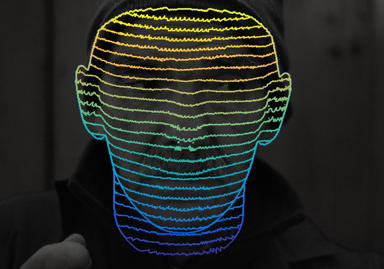}}\hspace{0.0005cm}
\subfloat[Estimated $V$ ($\hat{u}^v$)]{\includegraphics[width=0.13775\textwidth]{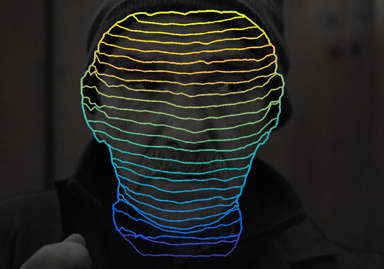}}\hspace{0.0005cm}
\subfloat[DenseReg]{\includegraphics[width=0.13775\textwidth]{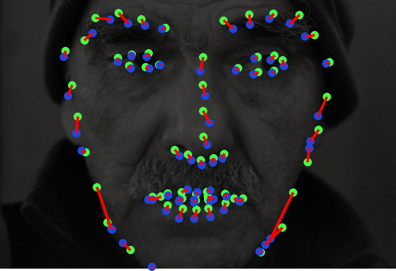}}\hspace{0.0005cm}
\subfloat[DenseReg+MDM]{\includegraphics[width=0.13775\textwidth]{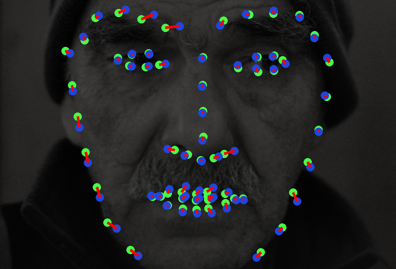}}\\
\vspace{-0.13cm}

\caption{Qualitative Results. Ground-truth and estimated deformation-free coordinates and  landmarks obtained from DenseReg and DenseReg+MDM are presented. Estimated landmarks(blue), ground-truth(green), lines between estimated and ground-truth landmarks(red).}
\label{fig:qualitative}
\vspace{-0.3cm}

\end{figure*}

Note that the closest deformation-free coordinates among all \emph{visible} pixels to a landmark point is not necessarily the correct corresponding landmark. This phenomenon is called ``landmark marching''~\cite{zhu2015high} and mostly affects the jaw landmarks which are dependent on changes in head pose. It should be noted that we do not use any explicit supervision for landmark detection nor focus on ad-hoc methods to cope with this issue. Errors on jaw landmarks due to invisible coordinates and improvements thanks to deformable models can be observed in Fig.~\ref{fig:qualitative}.

Herein, we evaluate the landmark localization performance of DenseReg as well as the performance obtained by employing DenseReg as an initialization for deformable models~\cite{papandreou2008adaptive,tzimiropoulos2014gauss,antonakos2015feature,trigeorgis2016mnemonic} trained for the specific task. In the second scenario, we provide a slightly improved initialization with very small computational cost by reconstructing the detected landmarks with a PCA shape model that is constructed from ground-truth annotations.

We present experimental results using the very challenging 300W benchmark. This is the testing database that was used in the 300W competition~\cite{sagonas_iccv_300w_2013,sagonas2016300} - the most important facial landmark localization challenge. The error is measured using the point-to-point RMS error normalized with the interocular distance and reported in the form of Cumulative Error Distribution (CED). Figure~\ref{fig:300w} (bottom) presents some self-evaluations in which we compare the quality of initialization for deformable modelling between DenseReg and two other standard face detection techniques (HOG-SVM~\cite{king2015max}, DPM~\cite{mathias2014face}). The employed deformable models are the popular generative approach of patch-based Active Appearance Models (AAM)~\cite{papandreou2008adaptive,tzimiropoulos2014gauss,antonakos2015feature}, as well as the current state-of-the-art approach of Mnemonic Descent Method (MDM)~\cite{trigeorgis2016mnemonic}. It is interesting to notice that the performance of DenseReg without any additional deformable model on top, already outperforms even HOG-SVM detection combined with MDM. Especially when DenseReg is combined with MDM, it greatly outperforms all other combinations.

\begin{table}[h]
\vspace{-0.05cm}

\centering
\scalebox{0.9}{
\begin{tabular}{|l|c|c|}
\hline
\emph{Method} & \emph{AUC} & \emph{Failure Rate (\%)}\\
\hline\hline
\textbf{DenseReg + MDM}                 & \textbf{0.5219} & \textbf{3.67} \\
DenseReg                                & 0.3605 & 10.83 \\
Fan et al.~\cite{fan2016approaching}    & 0.4802 & 14.83 \\
Deng et al.~\cite{deng2016m}            & 0.4752 & 5.5   \\
Martinez et al.~\cite{martinez20162}    & 0.3779 & 16.0  \\
Cech et al.~\cite{vcech2016multi}       & 0.2218 & 33.83 \\
Uricar et al.~\cite{uvrivcavr2016multi} & 0.2109 & 32.17 \\
\hline
\end{tabular}
}
\caption{Landmark localization results on the 300W testing dataset using 68 points. Accuracy is reported as the AUC and the Failure Rate.}
\label{tab:300w}
\vspace{-0.15cm}
\end{table}

Figure~\ref{fig:300w} (top) compares DenseReg+MDM with the results of the latest 300W competition~\cite{sagonas2016300}.

We greatly outperform all competitors by a large margin. It should be noted that the participants of the competition did not have any restrictions on the amount of training data employed and some of them are industrial companies (e.g. Fan et al.~\cite{fan2016approaching}), which further illustrates the effectiveness of our approach. Finally, Table~\ref{tab:300w} reports the area under the curve (AUC) of the CED curves, as well as the failure rate for a maximum RMS error of $0.1$. Apart from the accuracy improvement shown by the AUC, we believe that the reported failure rate of $3.67\%$ is remarkable and highlights the robustness of DenseReg.

\subsection{Deformable Tracking}
\label{sec:exp_deformable_tracking}

\begin{table}[h]
\centering
\scalebox{0.9}{
\begin{tabular}{|l|c|c|}
\hline
\emph{Method} & \emph{AUC} & \emph{Failure Rate (\%)}\\
\hline\hline
\textbf{DenseReg + MDM} & \textbf{0.5937} & \textbf{4.57} \\
DenseReg                                           & 0.4320 & 8.1   \\
Yang et al.~\cite{yang2015facial}                  & 0.5832 & 4.66  \\
Xiao et al.~\cite{xiao2015facial}                  & 0.5800 & 9.1   \\
Rajamanoharan et al.~\cite{rajamanoharan2015multi} & 0.5154 & 9.68  \\
Wu et al.~\cite{wu2015shape}                       & 0.4887 & 15.39 \\
Unicar et al.~\cite{uricar2015facial}              & 0.4059 & 16.7  \\
\hline
\end{tabular}
}
\caption{Deformable tracking results against the state-of-the-art on the 300VW testing dataset using 68 points. Accuracy is reported as AUC and the Failure Rate.}
\vspace{-0.25cm}
\label{tab:300vw}
\end{table}

For the challenging task of deformable face tracking on lengthy videos, we employ the testing database of the 300VW challenge~\cite{300VW,chrysos2015offline} - the only existing benchmark for deformable tracking ``in-the-wild''. The benchmark consists of $114$ videos ($\sim 218k$ frames in total) and includes videos captured in totally arbitrary conditions (severe occlusions and extreme illuminations).

The tracking is performed based on sparse landmark points, thus we follow the same strategy as in the case of landmark localization in Sec.~\ref{sec:exp_landmark_localization}.

We compare the output of DenseReg, as well as DenseReg+MDM which was the best performing combination for landmark localization in static images (Sec.~\ref{sec:exp_landmark_localization}), against the participants of the 300VW challenge.

Table~\ref{tab:300vw} reports the AUC and Failure Rate measures. DenseReg combined with MDM demonstrates better performance than the winner of the 300VW competition. It should be highlighted that our approach is not fine-tuned for the task-at-hand as opposed to the rest of the methods that were trained on video sequences and most of them make some kind of temporal modelling. Finally, similar to the 300W case, the participants were allowed to use unlimited training data (apart from the provided training seuqences), as opposed to DenseReg (and MDM) that were trained only on the $3148$ images mentioned in Sec.~\ref{sec:training_setup}. 
Please refer to the supplementary material for a more detailed presentation of the tracking results.

\subsection{Monocular Depth Estimation}
\label{sec:exp_depth}

\begin{figure}[!h]
\centering
\includegraphics[width=\linewidth]{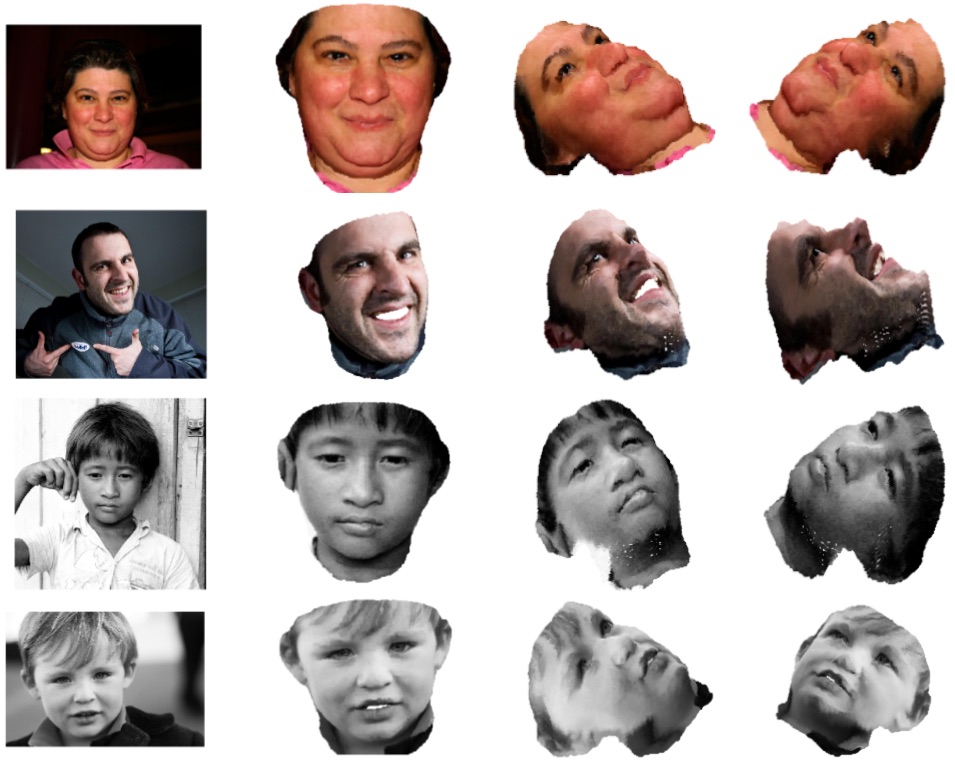}
\caption{Exemplar 3D renderings obtained using estimated depth values.}
\label{fig:depth}
\vspace{-0.35cm}
\end{figure}

The fitted template shapes also provide the depth from the image plane. We transfer this information to the visible pixels on the image using the same z-buffering operation used for the deformation-free coordinates (detailed in Sec.~\ref{sec:SDMs}  of the paper). We adopt this as an additional supervision signal: $Z \in [0,1]$ and add another branch to our network to estimate the depth along with the deformation-free coordinates. To our knowledge, there is no existing results in literature that would allow a quantitative comparison. We are 
providing example reconstructions using estimated monocular depth fields at Fig.\ref{fig:depth}.  We observe that this additional branch does not affect the performance of other branches and adds little to the complexity, since it is just a 1x1 convolution layer after the final shared convolutional layer. 

\subsection{Dense Correspondence for the Human Body}
\label{sec:exp_human}

\begin{figure}[!h]
\centering
\includegraphics[width=0.89\linewidth]{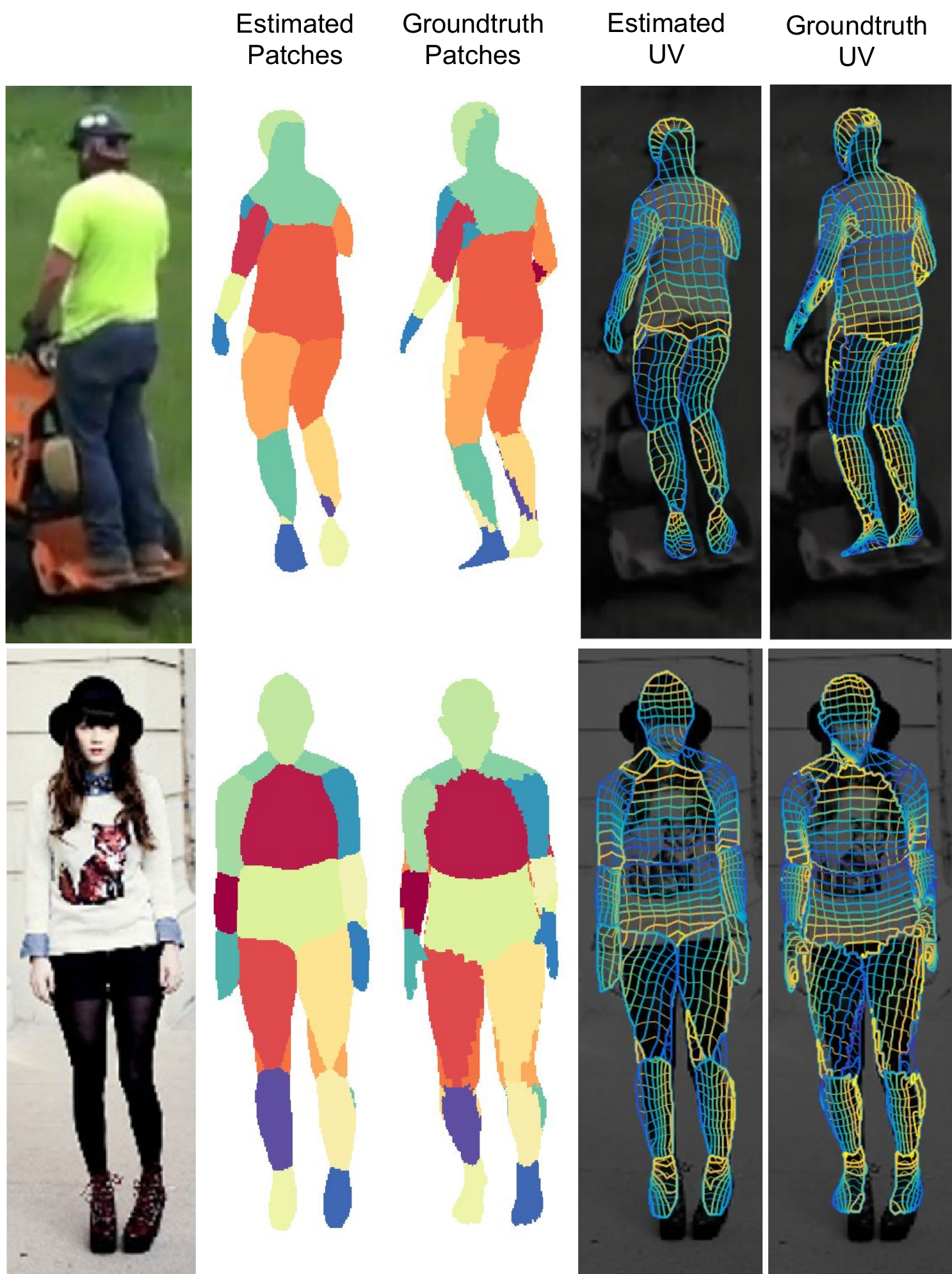}
\caption{Dense Correspondence for the human body.}
\vspace{-0.25cm}
\label{fig:DenseReg_Human}
\end{figure}

To portray that the DenseReg system can be used for articulated shapes of complex topology, we present results on the human shape. We use the recently proposed "Unite the People" (UP) dataset~\cite{lassner2017unite}, which provides a 3D deformable human shape model \cite{loper2015smpl} in correspondence with images from several publicly available datasets. We handle the complex geometry of the human shape by manually partitioning the surface into patches. We unwrap each patch using multidimensional scaling. The partitioning replaces the quantization and the rest of the system remains the same. Since there are no dense correspondence results between a 3D human model and image pixels in literature, we demonstrate the performance of our system through visual results from our test-set partition of the UP dataset in Fig.\ref{fig:DenseReg_Human}.
 
\subsection{Ear Landmark Localization}
\label{sec:exp_ear}



We have also performed experiments on the human ear. We employ the $602$ images and sparse landmark annotations that were generated in a semi-supervised manner by Zhou et al.~\cite{Zhou_2016_CVPR}. Due to the lack of a 3D model of the human ear, we apply Thin Plate Splines to bring the images into dense correspondence and obtain the deformation-free space. We perform landmark localization following the same procedure as in Sec.~\ref{sec:exp_landmark_localization}.
We split the images in $500$ for training and $102$ for testing.

\begin{figure}[h]
\centering
\includegraphics[width=\linewidth]{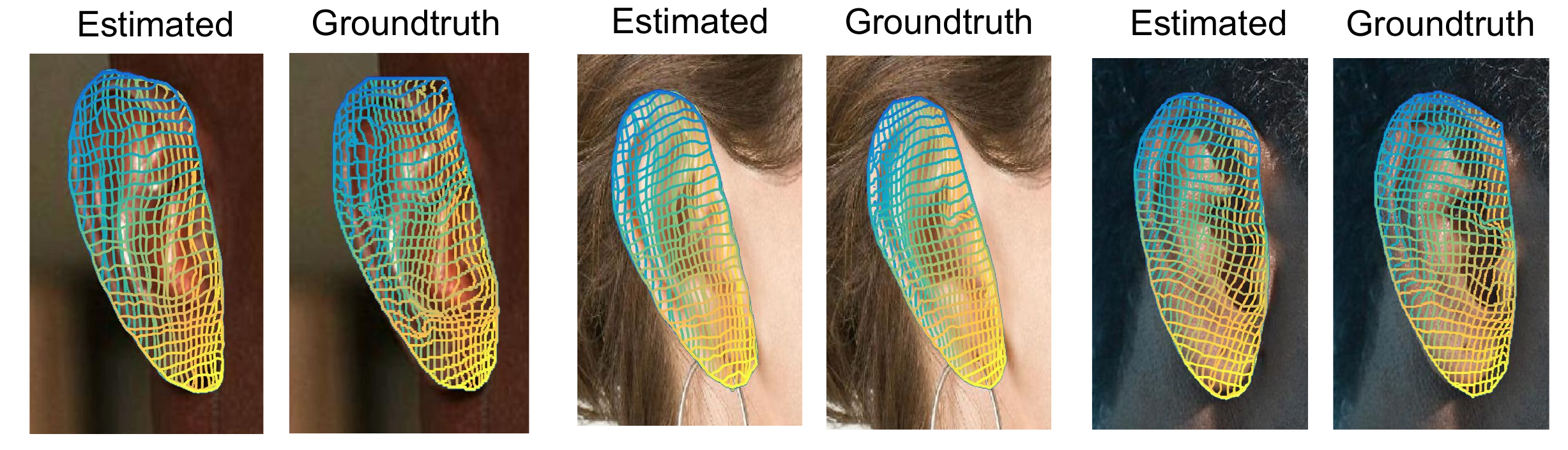}
\caption{Exemplar pairs of deformation-free coordinates of dense landmarks on human ear.}
\vspace{-0.5cm}
\label{fig:ears_examples}
\end{figure}

\begin{figure}[h!]
\centering
\includegraphics[width=\linewidth]{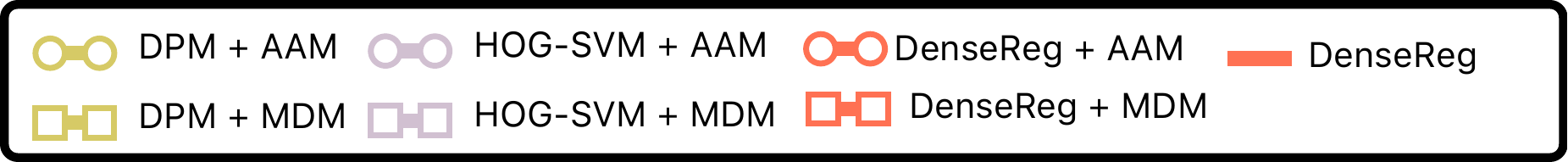}
\includegraphics[width=\linewidth]{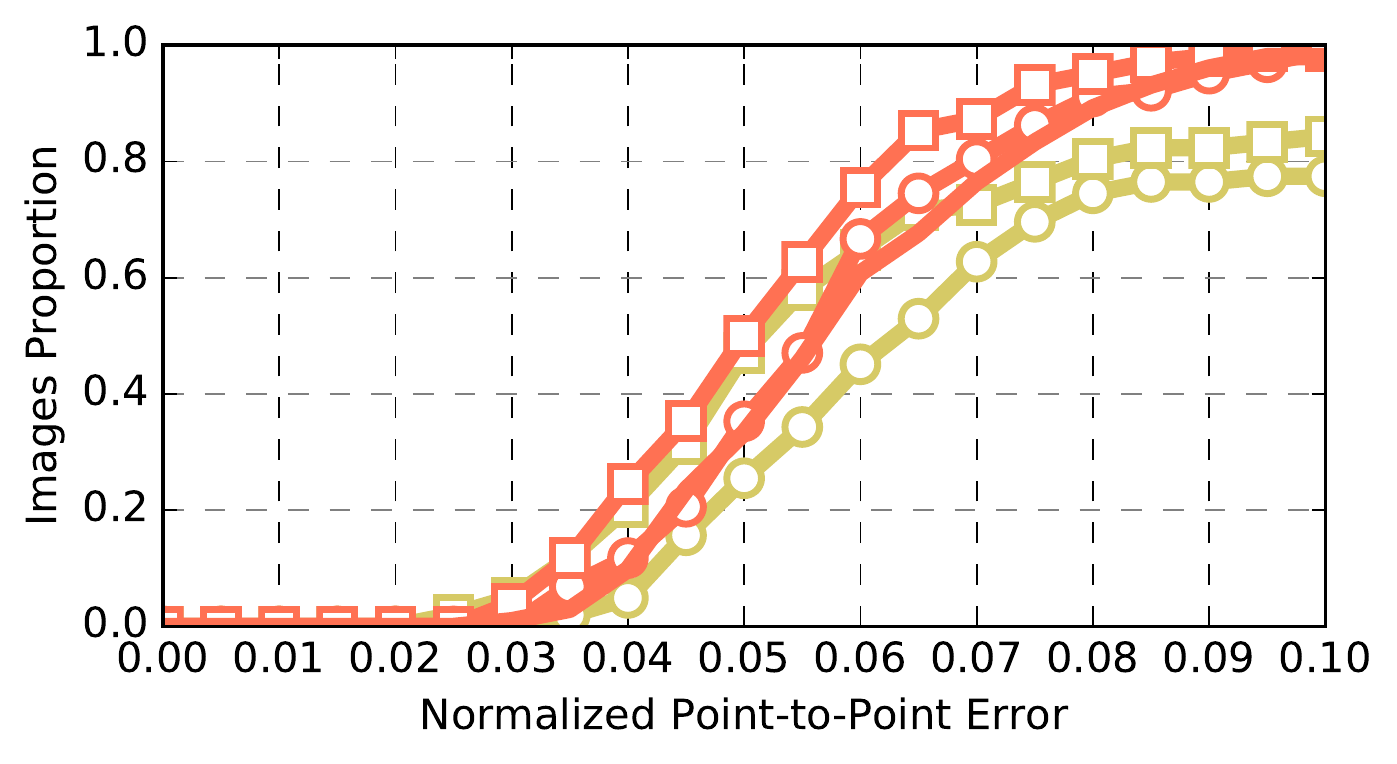}
\caption{Landmark localization results on human ear using 55 points. Accuracy is reported as Cumulative Error Distribution of normalized RMS point-to-point error.}
\vspace{-0.25cm}
\label{fig:ears}
\end{figure}

Given the lack of state-of-the-art deformable models on human ear, we compare DenseReg with DenseReg+AAM and DenseReg+MDM.  We also trained a DPM detector in order to compare the initialization quality with DenseReg. Figure~\ref{fig:ears} reports the CED curves based on the 55 landmark points using the RMS point-to-point error normalized by the bounding box average edge length. On Table.\ref{tab:ears}, we provide failure rate and the Area Under Curve(AUC) measures. Once again, the results are highly accurate even without improving DenseReg with a deformable model. We observe that DenseReg results are highly accurate and clearly outperforms the DPM based alternative even without a deformable model. Examples for dense human ear correspondence estimated by our system  are presented in Fig.~\ref{fig:ears_examples}.

\begin{table}[h!]
\centering
\begin{tabular}{|l|c|c|}
\hline
\emph{Method} & \emph{AUC} & \emph{Failure Rate (\%)}\\
\hline\hline
\textbf{DenseReg + MDM} & \textbf{0.4842} & \textbf{0.98} \\
DenseReg       &  0.4150 &   1.96 \\
DenseReg + AAM &  0.4263 &  0.98 \\
DPM + MDM      &  0.4160 &  15.69 \\
DPM + AAM      &  0.3283 &  22.55 \\
\hline
\end{tabular}
\caption{Landmark localization results on human ear using 55 points. Accuracy is reported as the Area Under the Curve (AUC) and the Failure Rate of the Cumulative Error Distribution of the normalized RMS point-to-point error.}
\label{tab:ears}
\end{table}

\section{Conclusion}
We propose a fully-convolutional regression approach for establishing dense correspondence fields between objects in natural images and three-dimensional object templates. We demonstrate that the correspondence information can successfully be utilised on problems that can be geometrically represented on the template shape. Throughout the paper, we focus on face shapes, where applications are abundant and benchmarks allow a fair comparison.  We show that using our dense regression method out-of-the-box  outperforms a state-of-the-art semantic segmentation approach for the task of face-part segmentation, while when used as an initialisation for SDMs,  we obtain the state-of-the-art results on the challenging 300W landmark localization challenge. We demonstrate the generality of our method by performing experiments on the human body and human ear shapes. We believe that our method will find ubiquitous use, since it can be readily used for face and human-body related tasks and can be easily integrated into many other correspondence problems.

{\small
\bibliographystyle{ieee}
\bibliography{DenseRegRefs}
}

\end{document}